\documentclass{article}

\usepackage{microtype}
\usepackage{graphicx}
\usepackage{subcaption}
\usepackage{booktabs} 
\usepackage{amsmath}
\usepackage{hyperref}

\usepackage{xcolor}
\usepackage{amsthm}
\usepackage{amssymb}

\usepackage{etoolbox}
\usepackage{multicol}

\newcommand{\eff}{\mathnormal{f}}

\usepackage[accepted]{icml2018}
\icmltitlerunning{Bucket Renormalization}

\begin{document}

\twocolumn[
\icmltitle{Bucket Renormalization for Approximate Inference}
\icmlsetsymbol{equal}{*}
\begin{icmlauthorlist}
\icmlauthor{Sungsoo Ahn}{kaist}
\icmlauthor{Michael Chertkov}{lanl,skol}
\icmlauthor{Adrian Weller}{cam,alan}
\icmlauthor{Jinwoo Shin}{kaist}
\end{icmlauthorlist}
\icmlaffiliation{kaist}{School of Electrical Engineering,
Korea Advanced Institute of Science and Technology, Daejeon,
Korea.}
\icmlaffiliation{lanl}{Theoretical Division, T-4 \& Center for Nonlinear Studies, Los Alamos National Laboratory, Los Alamos, NM 87545, USA}
\icmlaffiliation{skol}{Skolkovo Institute of Science and Technology, 143026 Moscow, Russia}
\icmlaffiliation{cam}{University of Cambridge, UK}
\icmlaffiliation{alan}{The Alan Turing Institute, UK}

\icmlcorrespondingauthor{Jinwoo Shin}{jinwoos@kaist.ac.kr}
\icmlkeywords{Machine Learning, ICML}
\vskip 0.3in
]

\printAffiliationsAndNotice{}

\begin{abstract}
Probabilistic graphical models are a key tool in machine learning applications. Computing the partition function, i.e., normalizing constant,
is a fundamental task of statistical inference but it is generally computationally intractable, leading to extensive study of
approximation methods. 
Iterative variational methods 
are a popular and successful family of approaches. 
However, even state of the art variational methods
can return poor results or fail to converge on difficult instances. 
In this paper,
we instead consider computing the partition function via sequential summation over 
variables. We develop robust approximate algorithms by combining ideas from
mini-bucket elimination 
with tensor network and renormalization group methods from statistical physics.
The resulting ``convergence-free'' methods show good
empirical performance on both synthetic and real-world benchmark models,
even for difficult instances.
\end{abstract}

\section{Introduction}
Graphical Models (GMs) express the 
factorization of the joint multivariate 
probability distribution over subsets of variables 
via graphical relations among them. 
They have played an important role in many 
fields, including 
computer vision \cite{freeman2000learning}, 
speech recognition \cite{bilmes2004graphical},
social science \cite{scott2017social} and 
deep learning \cite{hinton2006reducing}.
Given a GM, computing the partition function $Z$ (the normalizing constant) is the essence of other statistical inference tasks such as marginalization and sampling. 
The partition function can be 
calculated efficiently in tree-structured GMs through an iterative (dynamic programming) algorithm eliminating, i.e. summing up, variables sequentially. In principle, the elimination strategy extends to arbitrary loopy graphs, but the computational complexity is exponential in the tree-width, e.g.,
the junction-tree method \cite{shafer1990probability}.
Formally, the computation task is 
\#P-hard even to approximate \cite{jerrum1993polynomial}.

Variational approaches are often the most popular practical choice for approximate computation of the partition function. 
They map the counting problem into an approximate optimization problem stated over
a polynomial (in the graph size) number of variables. The optimization is typically solved iteratively via a message-passing algorithm, e.g.,
mean-field \cite{parisi1988statistical}, 
belief propagation \cite{pearl1982reverend}, 
tree-reweighted \cite{wainwright2005new}, 
 or 
gauges and/or re-parametrizations \cite{ahn2017gauge, ahn2018gauged}.
Lack of accuracy control and difficulty in forcing convergence in an acceptable number of steps are, unfortunately, typical for hard GM instances. 
Markov chain Monte Carlo methods (e.g., see  
\citealp{alpaydin2014introduction}) 
are also popular to approximate the partition function,
but typically suffer, even more than variational methods,  from slow convergence/mixing.

Approximate elimination is a sequential method to estimate the partition function.  Each step consists of summation over variables followed by (or combined with) approximation of the resulting complex factors. Notable flavors of this method include 
truncation of the Fourier coefficients \cite{xue2016variable}, approximation by random mixtures of rank-$1$ tensors \cite{wrigley2017tensor}, and arguably the most popular, 
elimination over mini-buckets \cite{dechter2003mini,liu2011bounding}. One advantage of the mini-bucket elimination approach is 
the ability to control the trade-off between computational complexity and approximation quality by
adjusting an induced-width parameter.  
Note that analogous control in variational methods, such as varying region sizes in generalized belief propagation  
\cite{yedidia2001generalized}, typically results in much more complicated optimization formulations to solve.
Another important advantage of mini-bucket elimination is that it is always guaranteed to terminate and, usually, it does so quickly. 
This is in contrast to iterative message-passing implementations of variational methods which can be notoriously slow on difficult instances. 

{\bf Contribution.} 
We improve the approximation quality of mini-bucket methods 
using tensor network and renormalization 
group approaches from statistical physics.
In this regard,  our method extends a series of recent papers 
exploring multi-linear tensor network transformations/contractions   
\cite{novikov2014putting, wrigley2017tensor, ahn2017gauge, ahn2018gauged}.
More generally, tensor network renormalization 
algorithms \cite{levin2007tensor, evenbly2015tensor} 
have been proposed in the quantum and statistical physics literature
for estimating partition functions. The algorithms consist of coarse-graining 
the graph/network by contracting sub-graphs/networks using a low-rank 
projection as a subroutine. 
However, 
the existing renormalization methods in the physics literature
have focused primarily on a 
restricted class of tensor-factorized models over regular grids/lattices,\footnote{
The special models are related to what may be called Forney-style grids/lattices \cite{forney2001codes} in the GM community.} while
factor-graph models \cite{clifford1990markov} over generic graphical structures
are needed in most machine learning applications.

For generalizing them to factor-graph models,
one would face at two challenges:
(a) coarse-graining of the tensor network 
relies on the periodic structure of  
grid/lattices 
and (b) its low-rank projections are only 
defined on ``edge variables'' 
that allows only two adjacent factors. 
To overcome them, 
we first replace the coarse-graining step by 
sequential elimination 
of the mini-bucket algorithms,
and then use the strategy of 
``variable splitting'' 
in order to generate 
auxiliary edge variables. 
Namely, we
combine ideas from tensor network renormalization and the mini-bucket schemes
where one is benefical to the other.
We propose two algorithms, which we call MBR and GBR: 
\begin{itemize}
\item \emph{Mini-bucket renormalization} (MBR) consists of sequentially splitting summation over the current (remaining) set of variables into subsets --
multiple mini-buckets which are then  ``renormalized''. We show that this process is, in fact, equivalent to applying low-rank projections on the mini-buckets to approximate the variable-elimination process,  thus resulting in better approximation than the original mini-bucket methods. In particular, we show how to resolve approximate renormalization locally and efficiently through application of truncated singular value decomposition (SVD) over small matrices.

\item 
While MBR is based on a sequence of local low-rank approximations applied to the mini-buckets, 
 \emph{global-bucket renormalization} (GBR) extends MBR by approximating mini-buckets globally. 
This is achieved by first applying MBR to mini-buckets, then calibrating the choice of low rank projections by minimizing the partition function 
approximation error with respect to renormalization of the ``global-bucket''. 
Hence, GBR takes additional time to run but may be expected to yield better accuracy.

\end{itemize}

Both algorithms are easily applicable to arbitrary GMs 
with interactions (factors) of high orders, hyper-graphs and large alphabets.
We perform extensive experiments 
on synthetic (Ising models on complete and grid graphs) and real-world models from the UAI dataset.
In our experiments, both MBR and GBR 
show performance 
superior to other state-of-the-art 
elimination and variational algorithms. 

\section{Preliminaries}
{\bf Graphical model.}
Consider a hyper-graph
$\mathcal{G} = (\mathcal{V}, \mathcal{E})$ with
vertices $\mathcal{V} = \{1,\cdots,n\}$ and
hyper-edges $\mathcal{E} \subset 2^{\mathcal V}$. A graphical 
model (GM) $\mathcal{M}= (\mathcal G, \mathcal{F})$ associates a 
collection of  $n$ discrete random variables
$\mathbf{x}=[x_{i}: i \in \mathcal{V}]\in\mathcal{X}_{\mathcal{V}}=\prod_{i\in\mathcal{V}}\mathcal{X}_{i}$
with the following joint probability distribution:
\begin{equation*}
\text{Pr}(\mathbf{x}) =
\frac{1}{Z}\prod_{\alpha\in\mathcal{E}}
\eff_{\alpha}(\mathbf{x}_{\alpha}),
\quad\quad
Z = \sum_{\mathbf{x}}
\prod_{\alpha\in\mathcal{E}}
\eff_{\alpha}(\mathbf{x}_{\alpha}),
\label{Forney}
\end{equation*}
where
$\mathcal{X}_{i}=\{1,2,\cdots d_i\}$,
$\mathbf{x}_{\alpha} = [x_{i}:i\in \mathbf{\alpha}]$,
$\mathcal{F} = \{\eff_{\alpha}\}_{\alpha \in \mathcal{E}}$
is a set of non-negative functions called {\it factors}, 
and $Z$ is the normalizing constant called the {\it partition function}
that is computationally intractable.

\begin{algorithm}[ht!]
\caption{Bucket Elimination (BE)}
\label{alg:bucket}
\begin{algorithmic}[1]
\STATE {\bf Input:} GM $\mathcal{M}^{\dagger}=(\mathcal{G}^{\dagger}, \mathcal{F}^{\dagger})$
and elimination order $o$.
\vspace{0.05in}
  \hrule
    \vspace{0.05in}
\STATE $\mathcal{F} \leftarrow \mathcal{F}^{\dagger}$
\FOR{$i$ in $o$}
\STATE $\mathcal{B}_{i} 
\leftarrow \{\eff_{\alpha}|\eff_{\alpha}\in \mathcal{F}, i \in \alpha \}$
\STATE Generate new factor $\eff_{\mathcal{B}_{i}\setminus \{i\}}$ by \eqref{eq:be}.
\STATE $\mathcal{F} \leftarrow \mathcal{F} \cup \{\eff_{\mathcal{B}_{i}\setminus \{i\}}\} \setminus \mathcal{B}_{i}$
\ENDFOR
\vspace{0.05in}
  \hrule
  \vspace{0.05in}
\STATE {\bf Output:} $Z = \prod_{\eff_{\alpha}\in \mathcal{F}}\eff_{\alpha}$
\end{algorithmic}
\end{algorithm}

{\bf Mini-bucket elimination.}
{\it Bucket (or variable) elimination} (BE,  \citealp{dechter1999bucket, koller2009probabilistic}) is
a procedure for computing the partition function exactly 
based on sequential elimination of variables. 
Without loss of generality, 
we assume through out the paper that the elimination order is fixed $o = [1,\cdots,n]$. 
BE groups factors by placing each 
factor $\eff_{\alpha}$ in the 
``bucket'' $\mathcal{B}_{i}\subset \mathcal F$ of its 
earliest argument 
$i\in \alpha$ 
appearing in the elimination order $o$.
Next, BE eliminates the variable by introducing a new factor
marginalizing the product of factors in it, 
i.e., 
\begin{equation}\label{eq:be}
\eff_{\mathcal{B}_{i}\setminus \{i\}}
(\mathbf{x}_{\mathcal{B}_{i}\setminus \{i\}})
= \sum_{x_{i}}
\prod_{\eff_{\alpha}\in \mathcal{B}_{i}}
\eff_{\alpha}(\mathbf{x}_{\alpha}).
\end{equation}
Here,
$\mathbf{x}_{\mathcal{B}_{i}\setminus \{i\}}$
abbreviates 
$\mathbf{x}_{\mathcal{V}(\mathcal{B}_{i})\setminus \{i\}}$,
where $\mathcal{V}(\mathcal{B}_{i})$
indicates the set of variables associated with the
bucket $\mathcal{B}_{i}$.
The subscript in
$\eff_{\mathcal{B}_{i}\setminus \{i\}}$
represents a similar abbreviation.
Finally, the new 
function $\eff_{\mathcal{B}_{i}\setminus \{i\}}$ 
is added to another bucket 
corresponding to its earliest argument in the 
elimination order. 
Formal description of BE is given
in Algorithm \ref{alg:bucket}.

One can easily check that BE 
applies a distributive property for computing $Z$ exactly:
groups of factors corresponding to buckets
are summed out sequentially,
and then the newly generated factor
(without the eliminated variable) is
added to another bucket. 
The computational cost of BE is
exponential with respect 
to the number of
uneliminated variables 
in the bucket, i.e., its complexity is 
$O\left(d^{\max_{i}|\mathcal{V}(\mathcal{B}_{i})|}|\mathcal{V}|\right).$
Here, 
$\max_{i\in\mathcal{V}}|\mathcal{V}(\mathcal{B}_{i})|$ is 
called the
{\it induced width} 
of $\mathcal{G}$
given the elimination order $o$, and the minimum possible
induced width across all possible $o$ 
is called the {\it tree-width}. 
Furthermore, 
we remark that 
summation of GM over variables defined on the subset of vertices $\alpha$, 
i.e., $\sum_{\mathbf{x}_{\alpha}}\prod_{\beta\in\mathcal{E}}\eff_{\beta}$, 
can also be computed via BE 
in 
$
O(d^{\max_{i}|\mathcal{V}(\mathcal{B}_{i})| + |\mathcal{V}\setminus\alpha|}|\mathcal{V}|)
$ time by summing out variables in 
elimination order 
$o_{\alpha}$ on $\alpha$.

{\it Mini-bucket elimination} (MBE,  \citealp{dechter2003mini})
achieves lower complexity by
approximating each step of BE by
splitting the computation
of each bucket into several
smaller ``mini-buckets''.
Formally, for each variable $i$ in the 
elimination order $o$, 
the bucket $\mathcal{B}_{i}$ 
is partitioned into
$m_{i}$ mini-buckets
$\{\mathcal{B}_{i}^{\ell}\}_{\ell=1}^{m_{i}}$
such that 
$\mathcal{B}_{i} = \bigcup_{\ell=1}^{m_{i}}\mathcal{B}_{i}^{\ell}$ and $\mathcal{B}_{i}^{\ell_{1}}\cap \mathcal{B}_{i}^{\ell_{2}}=\emptyset$
for any $\ell_{1},\ell_{2}$.
Next, MBE generates new factors differently from BE as follows:
\begin{align}
     \eff_{\mathcal{B}_{i}^{\ell}\setminus \{i\}}
     (\mathbf{x}_{\mathcal{B}_{i}^{\ell}\setminus \{i\}})
     =
     \max_{x_{i}}
     \prod_{\eff_{\alpha} \in \mathcal{B}^{\ell}_{i}}
     \eff_{\alpha}(\mathbf{x}_{\alpha}),\label{eq:mbemax}
\end{align}
for all $\ell = 1,\cdots, m_{i}-1$ and
\begin{align}
     \eff_{\mathcal{B}_{i}^{m_{i}}\setminus \{i\}}
     (\mathbf{x}_{\mathcal{B}_{i}^{m_{i}}\setminus \{i\}})
     =
     \sum_{x_{i}}
     \prod_{\eff_{\alpha} \in \mathcal{B}^{m_{i}}_{i}}
     \eff_{\alpha}(\mathbf{x}_{\alpha}).\label{eq:mbesum}
\end{align}
Other steps are equivalent to that of BE.
Observe that 
MBE 
replaces the exact marginalization of the bucket
in \eqref{eq:be}
by its upper bound,
i.e.,
$\sum_{x_{i}}
\prod_{\eff_{\alpha}\in\mathcal{B}_{i}}\eff_{\alpha}
\leq
\prod_{\ell=1}^{m}
\eff_{\mathcal{B}_{i}^{\ell}\setminus \{i\}}
$, 
and hence yields an
upper bound of $Z$. 
We remark that
one could instead obtain
a lower bound for $Z$
by replacing $\max$
by $\min$ in \eqref{eq:mbemax}.

Note that one has to choose mini-buckets for MBE carefully
as their sizes typically 
determine complexity and accuracy:
smaller mini-buckets may be better for speed,
but worse in accuracy. 
Accordingly, MBE has an additional 
induced width bound parameter 
$ibound$ as the maximal size of a mini-bucket, 
i.e.,
$|\mathcal{V}(\mathcal{B}_{i}^{\ell})|\leq
ibound + 1$. 
The time complexity of MBE is 
$O\left(d^{ibound+1}|\mathcal{E}|\cdot \max_{\alpha\in\mathcal{E}} |\alpha| \right), $
since the maximum number of mini-buckets is bounded by $|\mathcal{E}|\max_{\alpha\in\mathcal{E}} |\alpha|$.

\section{Mini-Bucket Renormalization}
\label{sec:rmbe}
We propose a new scheme, named 
mini-bucket renormalization
(MBR). Our approach 
approximates BE by 
splitting each 
bucket into 
several smaller 
mini-buckets to be 
``renormalized''. 
Inspired 
by tensor renormalization groups (TRG, \citealp{levin2007tensor}, see also references therein) in the physics literature, 
MBR utilizes low-rank approximations to
the mini-buckets instead of simply applying $\max$ (or $\min$) operations as in MBE. 
{Intuitively, MBR is similar to MBE 
but promises better accuracy.}

\begin{figure}[t!]
\vspace{-0.05in}
    \centering
    \begin{subfigure}[b]{0.99\linewidth}
    \centering
    \includegraphics[width=0.90\textwidth]{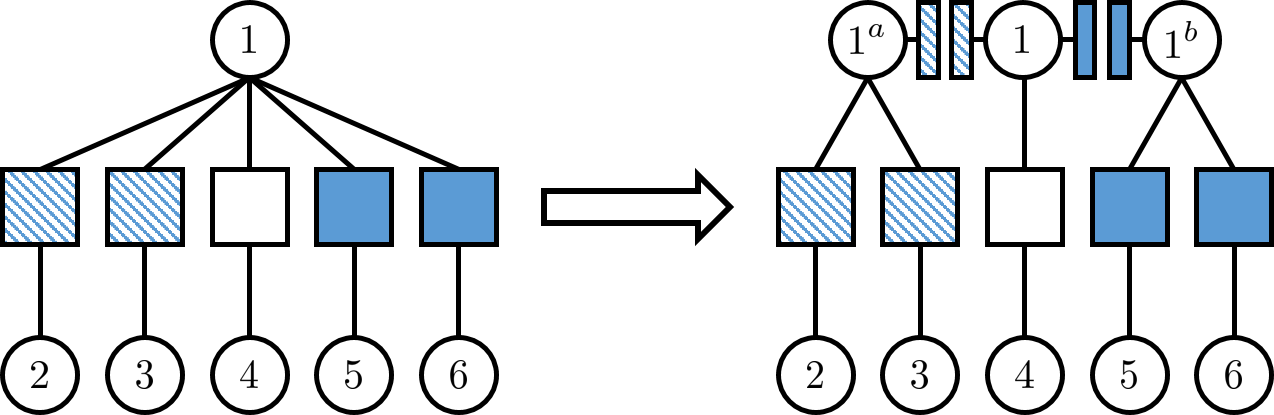}
    \end{subfigure}
    \caption{Renormalization process for mini-buckets $\mathcal{B}_{1}^{a} = \{\eff_{12},\eff_{13}\},
    \mathcal{B}_{1}^{b} = \{\eff_{15},\eff_{16}\}$ and 
    $\mathcal{B}_{1}^{c} = \{\eff_{14}\}$.
    }
    \label{fig:mbr}
    \vspace{-0.1in}
\end{figure}
\subsection{Algorithm Description}
For each variable $i$ in the elimination order $o$, 
MBR partitions a 
bucket $\mathcal{B}_{i}$ into
$m_{i}$ distinct mini-buckets 
$\{\mathcal{B}_{i}^{\ell}\}_{\ell=1}^{m_{i}}$
with maximal size bounded by $ibound$. 
Then for $\ell = 1,\cdots, m_{i}-1$, 
mini-bucket $\mathcal{B}_{i}^{\ell}$ 
is ``renormalized'' through  
replacing vertex $i$ 
by its replicate $i^{\ell}$ 
and then introducing local factors $r_{i}^{\ell}, r_{i^{\ell}}$ for error compensation, i.e., 
\begin{equation*}
    \widetilde{\mathcal{B}}_{i}^{\ell} 
    \leftarrow 
    \{\eff_{\alpha \setminus \{i\} \cup \{i^{\ell}\}}| 
    \eff_{\alpha}\in\mathcal{B}_{i}^{\ell}\}
    \cup \{r_{i}^{\ell}, r_{i^{\ell}}\}.
\end{equation*}
Here, $r_{i}^{\ell}, r_{i^{\ell}}$ 
are local 
``compensating/renormalizing 
factors'', 
chosen to approximate 
the factor  $\eff_{\mathcal{B}_{i}^{\ell}}=\prod_{\eff_{\alpha}\in\mathcal{B}_{i}^{\ell}}\eff_{\alpha}$
well, where MBE approximates it using \eqref{eq:mbemax} and \eqref{eq:mbesum}. 
{See Figure \ref{fig:mbr}
for illustration of the renormalization 
process.}
Specifically, the local compensating/renormalizing factors are chosen by 
solving the following optimization:
\begin{align}\label{eq:mbropt}
  \operatornamewithlimits{\mbox{min}}_{r_{i}^{\ell}, r_{i^{\ell}}}\qquad
    \sum_{\mathbf{x}_{\mathcal{B}^{\ell}_{i}}}
    \left(
    \eff_{\mathcal{B}_{i}^{\ell}}
    (\mathbf{x}_{\mathcal{B}_{i}^{\ell}})
    -
    \widetilde{\eff}_{\mathcal{B}_{i}^{\ell}}
    (\mathbf{x}_{\mathcal{B}_{i}^{\ell}})
    \right)^{2},
\end{align}
where $\widetilde{\eff}_{\mathcal{B}_{i}^{\ell}}$ is the
factor induced on $\mathbf{x}_{\mathcal{B}_{i}^{\ell}}$ 
from the renormalized mini-bucket $\widetilde{\mathcal{B}}_{i}^{\ell}$: 
\begin{align*}
    \widetilde{\eff}_{\mathcal{B}_{i}^{\ell}}(\mathbf{x}_{\mathcal{B}_{i}^{\ell}})&= 
    \sum_{x_{i}^{\ell}}\prod_{\eff_{\alpha} \in \widetilde{\mathcal{B}}_{i}^{\ell}}\eff_{\alpha}(\mathbf{x}_{\alpha})\\
    &=
    r_{i}^{\ell}(x_{i})
\sum_{x_{i^{\ell}}}
r_{i^{\ell}}(x_{i^{\ell}})
\prod_{\eff_{\alpha}
    \in \mathcal{B}_{i}^{\ell}}
    \eff_{\alpha}(x_{i^{\ell}}, \mathbf{x}_{\alpha\setminus \{i\} \cup \{i^{\ell}\}}).
\end{align*} 
We show that \eqref{eq:mbropt}
can be solved efficiently in Section \ref{subsec:complexity}. 
After all mini-buckets are processed, 
factors can be summed over 
the variables $x_{i^{1}}, \cdots, x_{i^{m_{i}-1}}$ and $x_{i}$ 
separately, i.e., 
introduce new factors as follows:
\begin{align}
    \eff_{\mathcal{B}_{i}^{\ell}\setminus \{i\}}
    (\mathbf{x}_{\mathcal{B}_{i}^{\ell}\setminus \{i\}}) 
    &=\sum_{x_{i^{\ell}}} 
    \prod_{\eff_{\alpha}\in 
    \widetilde{\mathcal{B}}_{i}^{\ell}\setminus \{r_{i}^{\ell}\}}
    \eff_{\alpha}(\mathbf{x}_{\alpha}), 
    \label{eq:mbr1}
    \\
    &= 
    \sum_{x_{i^{\ell}}}
    r^{\ell}_{i}(x_{i^{\ell}})
    \prod_{\eff_{\alpha}
    \in \mathcal{B}_{i}^{\ell}}
    \eff_{\alpha}(x_{i^{\ell}}, \mathbf{x}_{\alpha\setminus \{i\} \cup \{i^{\ell}\}}), \notag
\end{align} 
for $\ell = 1,\cdots, m_{i}-1$ 
and 
\begin{equation}\label{eq:mbr2}
\eff_{\mathcal{B}_{i}^{m_{i}}\setminus \{i\}}
    (\mathbf{x}_{\mathcal{B}_{i}^{m_{i}}\setminus \{i\}}) 
    = 
    \sum_{x_{i}}
    \prod_{\ell=1}^{m_{i}-1}r^{\ell}_{i}(x_{i^{\ell}})
    \prod_{\eff_{\alpha}
    \in \mathcal{B}_{i}^{m_{i}}}
    \eff_{\alpha}(\mathbf{x}_{\alpha}).
\end{equation}
Resulting factors are then added to 
its corresponding mini-bucket and 
repeat until all buckets are processed, like BE and MBE. 
Here, one can check that 
\begin{align*}
\prod_{\ell=1}^{m_{i}}
\eff_{\mathcal{B}_{i}^{\ell}\setminus \{i\}} 
(\mathbf{x}_{\mathcal{B}_{i}^{\ell}\setminus \{i\} })
&= 
\sum_{x_{i}}\eff_{\mathcal{B}_{i}^{m_{i}}}(\mathbf{x}_{\mathcal{B}_{i}^{m_{i}}})
\prod_{\ell=1}^{m_{i}-1}\widetilde{\eff}_{\mathcal{B}_{i}^{\ell}}(\mathbf{x}_{\mathcal{B}_{i}^{\ell}}),\\
&\approx
\sum_{x_{i}}
\prod_{\ell=1}^{m_{i}}\eff_{\mathcal{B}_{i}^{\ell}}(\mathbf{x}_{\mathcal{B}_{i}^{\ell}})
,
\end{align*} 
from \eqref{eq:mbropt} and MBR indeed approximates BE. 
The formal description of MBR is given in Algorithm \ref{alg:mbr}. 

\begin{algorithm}[ht!]
\caption{Mini-bucket renormalization (MBR)}
\label{alg:mbr}
\begin{algorithmic}[1]
\STATE {\bf Input:} GM $\mathcal M^{\dagger} = (\mathcal{G}^{\dagger}, \mathcal{F}^{\dagger})$, 
elimination order $o$ and induced width bound $ibound$.
\vspace{0.05in}
  \hrule
    \vspace{0.05in}
\STATE 
$\mathcal{F} \leftarrow \mathcal{F}^{\dagger}$
\FOR{$i$ in $o$}
\STATE $\mathcal{B}_{i} 
\leftarrow \{\eff_{\alpha}|\eff_{\alpha}\in \mathcal{F}, i \in \alpha \}$
\STATE Divide $\mathcal{B}_{i}$ into $m_{i}$ subgroups $\{\mathcal{B}_{i}^{\ell}\}_{\ell=1}^{m_{i}}$ such that 
$|\mathcal{V}(\mathcal{B}_{i}^{\ell})|\leq ibound +1$ for  $\ell=1,\cdots,m_{i}$. 
\FOR{$\ell=1,\cdots,m_{i}-1$}
\STATE Generate compensating factors $r_{i}^{\ell}, r_{i^{\ell}}$ 
by \eqref{eq:mbropt}.
\STATE Generate new factor $\eff_{\mathcal{B}_{i}^{\ell}\setminus\{i\}}$ 
by \eqref{eq:mbr1}.
\ENDFOR
\STATE Generate new factor $\eff_{\mathcal{B}_{i}^{m_{i}}\setminus \{i\}}$ by \eqref{eq:mbr2}.
\FOR{$\ell=1,\cdots, m_{i}$}
\STATE $\mathcal{F} \leftarrow \mathcal{F} \cup \{\eff_{\mathcal{B}_{i}^{\ell}\setminus\{i\}}\}\setminus \mathcal{B}_{i}^{\ell}$
\ENDFOR
\ENDFOR
\vspace{0.05in}
  \hrule
  \vspace{0.05in}
\STATE {\bf Output:} $Z=\prod_{\eff_{\alpha}\in \mathcal{F}}\eff_{\alpha}$
\end{algorithmic}
\end{algorithm}

\subsection{Complexity}
\label{subsec:complexity}
The optimization 
\eqref{eq:mbropt} is related to 
the rank-$1$ approximation on $\eff_{\mathcal{B}_{i}^{\ell}}$, 
which can be solved efficiently via (truncated) singular value decomposition (SVD). 
Specifically, let $\mathbf{M}$ be a $d \times d^{|\mathcal{V}(\mathcal{B}_{i}^{\ell})|-1}$ matrix representing
$\eff_{\mathcal{B}_{i}^{\ell}}$ 
as follows:
\begin{equation}\label{eq:svd}
    \mathbf{M}
    \left(
    x_{i},
    \sum_{j\in\mathcal{V}(\mathcal{B}_{i}^{\ell}), j \neq i}
    x_{j}
    \prod_{k \in \mathcal{V}(\mathcal{B}_{i}^{\ell}), k>j}d
    \right)
    =
    \eff_{\mathcal{B}_{i}^{\ell}}(\mathbf{x}_{\mathcal{B}_{i}^{\ell}}).
\end{equation}
Then rank-1 truncated SVD for $\mathbf{M}$
solves the following optimization: 
$$\min_{\mathbf{r}_{1}, \mathbf{r}_{2}}
\lVert
\mathbf{M} -
\mathbf{r}_{1}\mathbf{r}_{2}^{\top}
\mathbf{M}
\rVert_{F},$$
where optimization is over $d$-dimensional vectors $\mathbf{r}_{1}, \mathbf{r}_{2}$ and 
$\lVert\cdot\rVert_{F}$
denotes the Frobenious
norm. 
Namely, the solution $\mathbf{r}_{1}=\mathbf{r}_{2}$
becomes
the most significant (left) singular vector of $\mathbf{M}$, 
associated with the largest singular value.\footnote{
$\mathbf{r}_{1}=\mathbf{r}_{2}$ holds without forcing it.
} 
Especially, 
since $\mathbf{M}$ is a non-negative matrix, 
its most significant singular vector is 
always non-negative due to
the Perron-Frobenius
theorem \cite{perron1907theorie}. 
By letting $r_{i^{\ell}}(x) = \mathbf{r}_{1}(x)$ and $r_{i}^{\ell}(x) = \mathbf{r}_{2}(x)$, 
one can check that this 
optimization is 
equivalent to \eqref{eq:mbropt}, where
in fact, $r_{i^{\ell}}(x) =r_{i}^{\ell}(x)$, i.e., they share the same values.
Due to the above observations,
the complexity of \eqref{eq:mbropt} is 
$N_{\text{SVD}}(\mathbf{M})$ that denotes
the complexity of SVD for matrix $\mathbf{M}$.
Therefore, the overall complexity
becomes 
$$O\left(N_{SVD}(\mathbf{M}) \cdot T\right)=
O\left(N_{\text{SVD}}(\mathbf{M})\cdot
|\mathcal{E}|\cdot\max_{\alpha \in \mathcal{E}}|\alpha|\right),$$
where
$N_{\text{SVD}}(\mathbf{M}) = O(d^{ibound+2})$ in general, but typically
much faster in the existing SVD solver. 
\section{Global-Bucket Renormalization}\label{sec:global-bucket}
In the previous section, 
MBR  
only considers the local neighborhood
for 
renormalizing mini-buckets 
to approximate a single  marginalization 
process of BE. 
%
Here we extend the approach and propose 
global-bucket 
renormalization (GBR), which incorporates a global perspective. 
The new scheme re-updates the choice of
compensating local factors 
obtained in MBR 
by considering factors 
that were 
ignored during the original process. 
In particular, GBR directly 
minimizes 
the error in the 
partition function 
from each renormalization, 
aiming for improved accuracy compared to MBR. 
\subsection{Intuition and Key-Optimization}
\begin{figure*}[ht!]
\vspace{-0.05in}
\centering 
\begin{subfigure}[b]{0.99\linewidth}
    \centering
    \includegraphics[width=0.99\textwidth]{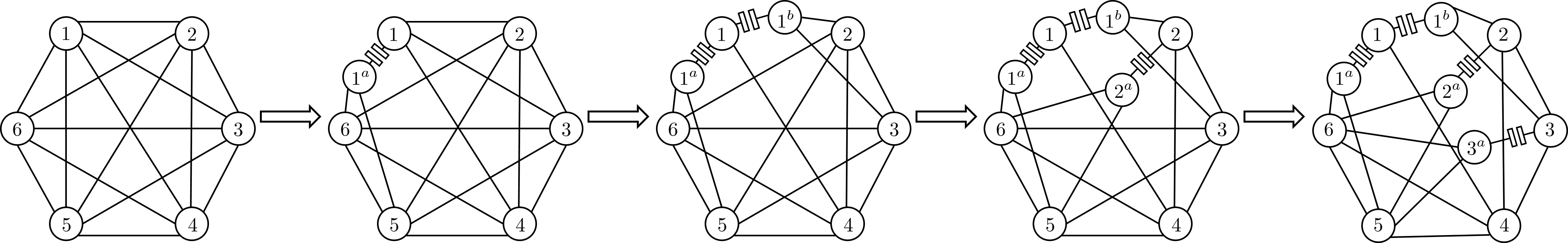}
    \end{subfigure}
    \caption{Example of GM renormalization on complete graph with 
    size $|\mathcal{V}|=6$ corresponding to execution of MBR with  elimination order $o=1,\cdots,6$ and $ibound=2$. 
    Here, pairwise factors between variables are assumed 
    to exist inside edges.
    Partition function of final GM is 
    able to be computed from BE with induced width $2$ and elimination order $\widetilde{o}=[1^{a},1^{b},1,2^{a},2,3^{a},3,4,5,6]$.}
    \label{fig:gm_renormalization}
    \vspace{-0.1in}
\end{figure*}

{\bf Renormalized GMs in MBR.} 
For providing the underlying design principles of GBR,
we first track 
an intermediate estimation 
of the partition function made during MBR by characterizing the corresponding sequence of 
renormalized GMs. 
Specifically, 
we aim for constructing a sequence of $T+1$ 
GMs 
$\mathcal{M}^{(1)},\cdots,\mathcal{M}^{(T+1)}$ with  
$T = \sum_{i=1}^{n}(m_{i}-1)$ 
by 
breaking each $i$-th iteration 
of MBR into $m_{i}-1$ steps of GM renormalizations,
$\mathcal{M}^{(1)}$ is the original GM, 
and each transition from $\mathcal{M}^{(t)}$ to $\mathcal{M}^{(t+1)}$ 
corresponds to 
renormalization of 
some mini-bucket $\mathcal{B}_{i}^{\ell}$ 
to $\widetilde{\mathcal{B}}_{i}^{\ell}$. 
Then,
the 
intermediate estimation for
the original partition function $Z$ at 
the $t$-th step 
is 
partition function $Z^{(t)}$ of $\mathcal{M}^{(t)}$ where 
the last one $Z^{(T+1)}$ is the output of MBR. 

To this end, 
we introduce 
``scope'' 
for each factor $\eff_{\alpha}$
appearing in MBR 
to indicate which parts of GM 
are renormalized at each step. 
In particular, the scope $\mathcal{S}_{\eff_{\alpha}} 
= (\mathcal{G}_{\eff_{\alpha}}, \mathcal{F}_{\eff_{\alpha}})$ consists of a graph $\mathcal{G}_{\eff_{\alpha}} 
= (\mathcal{V}_{\eff_{\alpha}}, 
\mathcal{E}_{\eff_{\alpha}})$ 
and set of factors $\mathcal{F}_{\eff_{\alpha}}$
that are associated to $\eff_{\alpha}$ as follows: 
\begin{equation*}
\eff_{\alpha}(\mathbf{x}_{\alpha}) 
= 
\sum_{\mathbf{x}_{\mathcal{V}_{\eff_{\alpha}}\setminus\alpha}}
\prod_{\eff_{\beta}\in\mathcal{F}_{\eff_{\alpha}}}
\eff_{\beta}(\mathbf{x}_{\beta}).  
\end{equation*}
Initially, scopes associated with initial factors  
are defined by themselves, 
i.e., 
\begin{equation}\label{eq:initscope}
\mathcal{S}_{\eff_{\alpha}} \leftarrow ((\alpha,\{\alpha\}), 
\{\eff_{\alpha}\}),
\end{equation}
for each factor 
$\eff_{\alpha}$ of $\mathcal{M}^{(1)}$, 
and others of $\mathcal{M}^{(t)}= ((\mathcal V^{(t)}, \mathcal E^{(t)}), \mathcal F^{(t)})$ with $t\geq 2$ are to be defined 
iteratively under the MBR process, as we describe in what follows.

Consider the $t$-th iteration of MBR, 
where mini-bucket $\mathcal{B}_{i}^{\ell}$ 
is being renormalized into 
$\widetilde{\mathcal{B}}_{i}^{\ell}$. 
Then scope $\mathcal{S}_{\eff}$ for all 
$\eff \in \mathcal{B}_{i}^{\ell}$ 
goes through renormalization by 
replacing every $i$ in the scope by $i^{\ell}$ 
as follows:
\begin{equation}\label{eq:factorrenorm}
    \widetilde{S}_{\eff}
    \leftarrow
    (\mathcal{V}_{\eff}\setminus \{i\} \cup \{i^{\ell}\}, 
    \{\bar{\alpha}| \alpha \in \mathcal{E}_{\eff}\}, 
    \{\eff_{\bar{\alpha}}|\eff_{\alpha} \in \mathcal{F}_{\eff}\}),
\end{equation}
where 
$\bar{\alpha}=\begin{cases}\alpha\setminus \{i\} \cup \{i^{\ell}\}&\mbox{ 
if}~i\in\alpha\\\qquad \alpha&\mbox{otherwise}\end{cases}$.
Then, the respective GM is 
renormalized accordingly for the change of scopes, 
in addition to compensating factors $r_{i}^{\ell}, r_{i^{\ell}}$: 
\begin{align}
\mathcal{V}^{(t+1)} &\leftarrow \mathcal{V}^{(t)}\cup \{i^{\ell}\},\notag\\
\mathcal{E}^{(t+1)} &\leftarrow \mathcal{E}^{(t)} 
\setminus\mathcal{E}_{\mathcal{B}_{i}^{\ell}}
\cup \widetilde{\mathcal{E}}_{\mathcal{B}_{i}^{\ell}}
\cup \{\{i\}, \{i^{\ell}\}\} 
,\label{eq:gmrenorm}\\
\mathcal{F}^{(t+1)} &\leftarrow \mathcal{F}^{(t)} 
\setminus\mathcal{F}_{\mathcal{B}_{i}^{\ell}}
\cup \widetilde{\mathcal{F}}_{\mathcal{B}_{i}^{\ell}}
\cup \{r_{i}^{\ell}, r_{i^{\ell}}\}, \notag
\end{align}
where 
$\mathcal{E}_{\mathcal{B}_{i}^{\ell}} = \cup_{\eff \in \mathcal{B}_{i}^{\ell}} \mathcal{E}_{\eff}, 
$ and other union of scope components 
$\widetilde{\mathcal{E}}_{\mathcal{B}_{i}^{\ell}}, 
\mathcal{V}_{\mathcal{B}_{i}^{\ell}}, 
\widetilde{\mathcal{V}}_{\mathcal{B}_{i}^{\ell}}, 
\mathcal{F}_{\mathcal{B}_{i}^{\ell}},
\widetilde{\mathcal{F}}_{\mathcal{B}_{i}^{\ell}}$ are 
defined similarly.
Finally, 
scope 
$\mathcal{S}_{\eff_{\mathcal{B}_{i}^{\ell}\setminus \{i\}}}$ 
for newly generated factors 
$\eff_{\mathcal{B}_{i}^{\ell}\setminus \{i\}}$
is
\begin{equation}\label{eq:factorscope1}
\mathcal{S}_{\eff_{\mathcal{B}_{i}^{\ell}\setminus \{i\}}}
\leftarrow
((
\widetilde{\mathcal{V}}_{\mathcal{B}_{i}^{\ell}},
\widetilde{\mathcal{E}}_{\mathcal{B}_{i}^{\ell}}
\cup \{\{i^{\ell}\}\}
),
\widetilde{\mathcal{F}}_{\mathcal{B}_{i}^{\ell}}
\cup \{r_{i^{\ell}}\}
).
\end{equation}
Furthermore, if $\ell = m_{i}-1$, we have
\begin{equation}\label{eq:factorscope2}
\mathcal{S}_{\eff_{\mathcal{B}_{i}^{m_{i}}\setminus \{i\}}}
\leftarrow (
(
\mathcal{V}_{\mathcal{B}_{i}^{m_{i}}}, 
\mathcal{E}_{\mathcal{B}_{i}^{m_{i}}}
\cup \{\{i\}\}
), 
\mathcal{F}_{\mathcal{B}_{i}^{m_{i}}}
\cup \{r_{i}^{\ell}\}_{\ell=1}^{m_{i}-1}
). 
\end{equation} 
This is repeated until 
the MBR process terminates, as formally described in
Algorithm \ref{alg:mbr_renormalization}. 
By construction, 
the output of MBR is equal to
the partition function of the last GM 
$\mathcal{M}^{(T+1)}$,
which is computable via BE 
with induced width smaller than $ibound + 1$ 
given elimination order 
\begin{equation}
\label{eq:elimination-order}
\widetilde{o} = 
[1^{1},\cdots,1^{m_{1}-1}, 1, \cdots, n^{1},\cdots, n^{m_{n}-1}, n].
\end{equation}
See Algorithm \ref{alg:mbr_renormalization} for the formal description of this process,
and Figure \ref{fig:gm_renormalization} for an example.

{\bf Optimizing intermediate approximations.}
Finally, we provide 
an explicit optimization formulation
for minimizing the 
change of 
intermediate 
partition functions 
in terms of induced factors.
Specifically, 
for each $t$-th renormalization, i.e., 
from $\mathcal{M}^{(t)}$ 
to $\mathcal{M}^{(t+1)}$, 
we consider change of the following 
factor 
$\eff_{i}$ 
induced from global-bucket 
$
\mathcal{F}^{(t)}$ 
to variable $x_{i}$ in a 
``skewed'' manner as follows: 
\begin{align*}
    \eff_{i}(x_{i}^{(1)}, x_{i}^{(2)}) 
    :=&  \sum_{\mathbf{x}_{\mathcal{V}^{(t)}\setminus \{i\}}}
    \prod_{\eff_{\alpha}\in\mathcal{F}_{\mathcal{B}_{i}^{\ell}}}
    \eff_{\alpha}(x_{i}^{(1)}, \mathbf{x}_{\alpha\setminus \{i\}})\\
    &\cdot\prod_{\eff_{\alpha} \in \mathcal{F}^{(t)} \setminus \mathcal{F}_{\mathcal{B}_{i}^{\ell}}}
    \eff_{\alpha}(x_{i}^{(2)}, \mathbf{x}_{\alpha\setminus\{i\}}), 
\end{align*} 
where $x_{i}^{(1)},x_{i}^{(2)}$ are the 
newly introduced ``split variables'' that 
are associated with the same vertex $i$, but 
allowed to have different values 
for our purpose.
Next, the bucket $\mathcal{F}^{(t)}$ 
is renormalized into 
$\mathcal{F}^{(t+1)}$, 
leading to the induced factor of 
$\widetilde{\eff}_{i}$ 
defined as follows:
\begin{align*}
\widetilde{\eff}_{i}(x_{i}^{(1)}, x_{i}^{(2)}) 
:=& \sum_{\mathbf{x}_{\mathcal{V}^{(t+1)}\setminus \{i\}}}
    \prod_{\eff_{\alpha}\in\widetilde{\mathcal{F}}_{\mathcal{B}_{i}^{\ell}}\cup\{r_{i}^{\ell}, r_{i^{\ell}}\}}
    \eff_{\alpha}(x_{i}^{(1)}, \mathbf{x}_{\alpha\setminus \{i\}})
    \notag\\
    &\cdot \prod_{\eff_{\alpha} \in \mathcal{F}^{(t+1)} 
    \setminus \widetilde{\mathcal{F}}_{\mathcal{B}_{i}^{\ell}}\setminus \{r_{i}^{\ell}, r_{i^{\ell}}\}}
    \eff_{\alpha}(x_{i}^{(2)}, 
    \mathbf{x}_{\alpha\backslash \{i\}})\notag\\
    =&
    r_{i}^{\ell}(x_{i}^{(1)})
    \sum_{x_{i^{\ell}}}
    r_{i^{\ell}}(x_{i^{\ell}})
    \eff_{i}(x_{i^{\ell}}, x_{i}^{(2)}).
\end{align*}
Then change in $\eff_{i}$ 
is directly related with 
change in partition function since
$Z^{(t-1)}$ and $Z^{(t)}$ can 
be described as follows:
\begin{equation*}
    Z^{(t-1)} = \sum_{x_{i}}\eff_{i}(x_{i}, x_{i}), \quad
    Z^{(t)} = \sum_{x_{i}}\widetilde{\eff}_{i}(x_{i},x_{i}).
\end{equation*}
Consequently, 
GBR chooses to 
minimize the change in $\eff_{i}$ 
by re-updating 
$r_{i}^{\ell}, r_{i^{\ell}}$, i.e., 
it solves
\begin{equation}\label{eq:lbropt}
\min_{r_{i}^{\ell},r_{i^{\ell}}}
\sum_{x_{i}^{(1)},x_{i}^{(2)}} 
\bigg(
\eff_{i}(x_{i}^{(1)},x_{i}^{(2)}) 
- 
\widetilde{\eff}_{i}(x_{i}^{(1)},x_{i}^{(2)})
\bigg)^{2}.
\end{equation}
However, we remark that \eqref{eq:lbropt} 
is intractable since 
its objective is ``global'', 
and requires summation over 
all variables except one, i.e.,  
$\mathbf{x}_{\mathcal{V}^{(t)}\setminus \{i\}}$, 
and this is the key difference from \eqref{eq:mbropt} 
which seeks to minimize 
the error described by the local mini-bucket.
GBR avoids this issue by 
substituting $\eff_{i}$ by its 
tractable approximation $\mathnormal{g}_{i}$, 
which is to be described 
in the following section.
\begin{algorithm}[ht!]
\caption{GM renormalization}
\label{alg:mbr_renormalization}
\begin{algorithmic}[1]
\STATE {\bf Input:} GM $\mathcal M^{\dagger} = (\mathcal{G}^{\dagger}, \mathcal{F}^{\dagger})$, 
elimination order $o$ and induced width bound $ibound$.
\vspace{0.05in}
  \hrule
    \vspace{0.05in}
\STATE $\mathcal{M}^{(1)}\leftarrow \mathcal{M}^{\dagger}$
\STATE Run Algorithm \ref{alg:mbr} with 
input $\mathcal M^{(1)}, o, ibound$ to obtain mini-buckets  $\mathcal{B}_{i}^{\ell}$ and compensating factors $r_{i}^{\ell}, r_{i^{\ell}}$ for $i=1,\cdots, n$ and $\ell = 1,\cdots,m_{i}$.
\FOR{$\eff\in\mathcal{F}^{(1)}$}
\STATE Assign scope $\mathcal{S}_{\eff}$ for $\eff$ by \eqref{eq:initscope}. 
\ENDFOR
\FOR{$i$ in $o$}
\FOR{$\ell = 1,\cdots, m_{i}-1$}
\FOR{$\eff \in \mathcal{B}_{i}^{\ell}$}
\STATE Renormalize scope $\mathcal{S}_{\eff}$ for $\eff$ into $\widetilde{\mathcal{S}}_{\eff}$ by \eqref{eq:factorrenorm}.
\ENDFOR
\STATE Set $t = \sum_{j=1}^{i-1}(m_{i}-1) + \ell$. 
\STATE Renormalize GM $\mathcal{M}^{(t)}$ into $\mathcal{M}^{(t+1)}$ by \eqref{eq:gmrenorm}.
\STATE Assign scope 
$\mathcal{S}_{\eff_{\mathcal{B}_{i}^{\ell}\setminus \{i\}}}$ for factor $\eff_{\mathcal{B}_{i}^{\ell}\setminus \{i\}}$ 
by \eqref{eq:factorscope1}.
\ENDFOR
\STATE Assign scope 
$\mathcal{S}_{\eff_{\mathcal{B}_{i}^{m_{i}}\setminus \{i\}}}$ 
for factor $\eff_{\mathcal{B}_{i}^{m_{i}}\setminus \{i\}}$ 
by \eqref{eq:factorscope2}.
\ENDFOR
\vspace{0.05in}
  \hrule
  \vspace{0.05in}
\STATE {\bf Output:} Final GM $\mathcal M^{(T+1)}$ with $T = \sum_{i=1}^{n}(m_{i}-1)$.
\end{algorithmic}
\end{algorithm}

\subsection{Algorithm Description}
In this section, we provide a formal description of GBR.
First, consider the 
sequence of GMs  
$\mathcal{M}^{(1)}, \cdots, \mathcal{M}^{(T+1)}$ 
from interpreting 
MBR as GM renormalizations. 
One can observe that this 
corresponds to 
$T$ choices of 
compensating 
factors made at each renormalization, 
i.e., $r^{(1)},\cdots, r^{(T)}$ 
where $r^{(t)}(x) = r_{i}^{\ell}(x) = r_{i^{\ell}}(x)$ 
for the associated 
replicate vertex $i^{\ell}$.
GBR modifies this sequence iteratively 
by replacing intermediate choice of 
compensation $r^{(t)}$ 
by another choice $s^{(t)}(x)=s_{i}^{\ell}(x)=s_{i^{\ell}}(x)$ 
in reversed order, 
approximately 
solving \eqref{eq:lbropt} until 
all compensating factors are updated. 
Then, GBR outputs partition function $Z^{(T+1)}$ 
for $\mathcal{M}^{(T+1)}$ as an approximation of the 
original partition function $Z$.

Now we describe the process of choosing 
new compensating factors 
$s_{i}^{\ell}, s_{i^{\ell}}$ 
at $t^{\prime}$-th iteration of GBR 
by approximately 
solving \eqref{eq:lbropt}. 
To this end, the $t^{\prime}$-th choice 
of compensating factors are 
expressed as follows:
\begin{equation}\label{eq:lbr1}
r^{(1)},\cdots, r^{(t)},s^{(t+1)},\cdots, s^{(T)},
\end{equation}
with $t = T- t^{\prime} + 1$ and
$s^{(t+1)},\cdots,s^{(T)}$ 
already chosen in the 
previous iteration of GBR. 
Next, 
consider sequence of GMs 
$\widehat{\mathcal M}^{(1)}, \cdots, \widehat{\mathcal M}^{(T+1)}$ 
that were generated similarly 
with GM renormalization corresponding to MBR, 
but with compensating factors chosen 
by \eqref{eq:lbr1}. 
Observe that the first $t$ 
renormalizations of GM 
correspond 
to those of MBR 
since the updates are done in reverse order, 
i.e., 
$\widehat{\mathcal M}^{(t^{\prime})} = \mathcal M^{(t^{\prime})}$ for $t^{\prime} < t + 1$. 
Next, 
$\eff_{i}$ in 
\eqref{eq:mbropt} 
is expressed as 
summation over 
$\mathbf{x}
_{\widehat{\mathcal{V}}^{(t+1)}\setminus \{i, i^{\ell}\}}
^{\ell}$ 
in $\widehat{\mathcal M}^{(t+1)}$, 
defined as follows: 
\begin{equation*}
\eff_{i}
(
x_{i^{\ell}}, 
x_{i}
) 
= 
\sum_{\mathbf{x}_{\widehat{\mathcal{V}}^{(t+1)}\setminus \{i, i^{\ell}\}}}
\prod_{\eff_{\alpha} \in \widehat{\mathcal{F}}^{(t+1)} \setminus \{r_{i}^{\ell} , r_{i^{\ell}}\}}
\eff_{\alpha}(\mathbf{x}_{\alpha}).
\end{equation*}
Since $\eff_{i}$ 
resembles the partition function 
in a way that it is also a summation  
over GM with small 
change in a set of factors, i.e., 
excluding local factors 
$r_{i}^{\ell}, r_{i^{\ell}}$, 
we expect 
$\eff_{i}$ to be 
approximated well by 
a summation $g_i$ over 
$\mathbf{x}_{\widehat{\mathcal{V}}^{(T+1)}\setminus \{i, i^{\ell}\}}$ 
in $\widehat{\mathcal M}^{(T+1)}$: 
\begin{equation*}
g_{i}(x_{i^{\ell}}, x_{i}) 
:= 
\sum_{\mathbf{x}_{\widehat{\mathcal{V}}^{(T+1)} 
\setminus \{i,i^{\ell}\}}}
\prod_{\eff_{\alpha} \in \widehat{\mathcal{F}}^{(T+1)} 
\setminus \{r_{i}^{\ell}, r_{i^{\ell}}\}}
\eff_{\alpha}(\mathbf{x}_{\alpha}), 
\end{equation*} 
which can be computed in 
$O(|\mathcal{E}|d^{ibound+2})$ complexity 
via appying BE in $\widehat{\mathcal M}^{(T+1)}$ 
with elimination order 
$\widetilde{o}\setminus \{i,i^{\ell}\}$ as in 
\eqref{eq:elimination-order}.
Then 
the following optimization is obtained as an  
approximation of \eqref{eq:lbropt}:
\begin{equation}\label{eq:lbropt2}
    \min_{s_{i}^{\ell}, s_{i^{\ell}}} \sum_{x_{i}^{(1)},x_{i}^{(2)}}
    \bigg(
    g_{i}(x_{i}^{(1)},x_{i}^{(2)}) 
    -
    \widetilde{g}_{i}(x_{i}^{(1)},x_{i}^{(2)}) 
    \bigg)^{2},
\end{equation}
where $\widetilde{g}_{i}$ corresponds to renormalized factor $\widetilde{\eff}_{i}$: 
\begin{equation*}
\widetilde{g}_{i}(x_{i^{\ell}}, x_{i}) 
    :=  
    s_{i}^{\ell}(x_{i^{\ell}})
    \sum_{x_{i^{\ell}}}
    s_{i^{\ell}}(x_{i^{\ell}})
    g_{i}(x_{i^{\ell}}, x_{i}).
\end{equation*}
As a result, 
one can expect choosing compensating factors from  \eqref{eq:lbropt2} 
to improve over that of 
\eqref{eq:mbropt} as long as 
MBR provides reasonable 
approximation for $\eff_{i}$.
The optimization is again solvable via rank-$1$ 
truncated SVD 
and the overall complexity of GBR
is 
\begin{equation*}
O\left(d^{ibound + 2} N_{\text{SVD}}(\mathbf{M}^{\text{global}}) \cdot|\mathcal{E}|^{2}\cdot \max_{\alpha \in \mathcal{E}}|\alpha|^{2}\right), 
\end{equation*}
where $N_{\text{SVD}}(\mathbf{M}^{\text{global}}) = O(d^{3})$ 
is the complexity for performing SVD on 
$d\times d$ matrix $\mathbf{M}^{\text{global}}$
representing function $g$ as in \eqref{eq:svd}.
While the formal description of GBR is conceptually 
a bit complicated, one can implement it efficiently.
Specifically, during the GBR process, 
it suffices to keep only the 
description of 
renormalized GM 
$\mathcal{M}^{(T+1)}$ 
with an ongoing set of 
compensating factors, 
e.g., \eqref{eq:lbr1}, 
in order to update compensating factors iteratively 
by \eqref{eq:lbropt2}. 
The formal description of  the GBR scheme 
is provided in 
Algorithm \ref{alg:lbr}. 
\begin{algorithm}[ht!]
\caption{Global-Bucket Renormalization (GBR)}
\label{alg:lbr}
\begin{algorithmic}[1]
\STATE {\bf Input:} GM $\mathcal{M}^{\dagger} = (G^{\dagger}, \mathcal{F}^{\dagger})$, 
elimination order $o$ and induced width bound $ibound$.
\vspace{0.05in}
  \hrule
    \vspace{0.05in}
\STATE Run Algorithm \ref{alg:mbr_renormalization} 
with input $\mathcal{M}^{\dagger}, o, ibound$ to 
obtain renormalized GM $\mathcal{M}$ 
and compensating factors $r_{i}^{\ell}$ 
for $i=1,\cdots, n$ and $\ell = 1,\cdots,m_{i}$.
\STATE Set the renormalized elimination order as follows:
\begin{equation*}
\widetilde{o} = 
[1^{1},\cdots,1^{m_{1}-1}, 1, \cdots, n^{1},\cdots, n^{m_{n}-1}, n].
\end{equation*}
\FOR{$i^{\ell}= n^{m_{n}-1},\cdots,n^{1},\cdots, 1^{m_{1}-1},\cdots,1^{1}$}
\STATE Generate $s_{i}^{\ell}, s_{i}^{\ell}$ by solving 
\begin{align*}
    \min_{s_{i}^{\ell}, s_{i^{\ell}}}& \sum_{x_{i}^{(1)},x_{i}^{(2)}}
    \bigg(
    g_{i}(x_{i}^{(1)}, x_{i}^{(2)}) 
    -
    \widetilde{g}_{i}(x_{i}^{(1)}, x_{i}^{(2)}) 
    \bigg)^{2},
\end{align*}
\STATE where $g_{i}, \widetilde{g}_{i}$ is defined as follows:
\begin{align*}
g_{i}(x_{i^{\ell}}, x_{i}) 
&= 
\sum_{\mathbf{x}_{\mathcal{V}\setminus \{i,i^{\ell}\}}}
\prod_{\eff_{\alpha} \in \mathcal{F}\setminus \{r_{i}^{\ell}, r_{i^{\ell}}\}}\eff_{\alpha}(\mathbf{x}_{\alpha}), \\
\widetilde{g}_{i}(x_{i^{\ell}}, x_{i}) 
    &=  
    s_{i}^{\ell}(x_{i^{\ell}})
    \sum_{x_{i}^{\ell}}
    s_{i^{\ell}}(x_{i}^{\ell})
    g_{i}(x_{i}^{\ell}, x_{i}), 
\end{align*}
with its computation done by BE with elimination order of 
$\widetilde{o}\setminus \{i, i^{\ell}\}$.
\STATE Update GM $\mathcal{M}$ by $\mathcal{F}\leftarrow\mathcal{F}\setminus \{r_{i}^{\ell}, r_{i^{\ell}}\} \cup \{s_{i}^{\ell}, 
s_{i^{\ell}}\}$.
\ENDFOR
\STATE Get $Z = \sum_{\mathbf{x}}\prod_{\eff_{\alpha}\in \mathcal{F}}\eff_{\alpha}(\mathbf{x}_{\alpha})$ 
via BE with elimination order $\widetilde{o}$.
\vspace{0.05in}
  \hrule
  \vspace{0.05in}
\STATE {\bf Output:} $Z$
\end{algorithmic}
\end{algorithm}
\vspace{-0.1in}

\section{Experimental Results}
\label{sec:experiment}
\begin{figure*}[ht!]
\vspace{-0.05in}
\begin{subfigure}[b]{0.245\linewidth}
    \centering
    \includegraphics[width=0.99\textwidth]{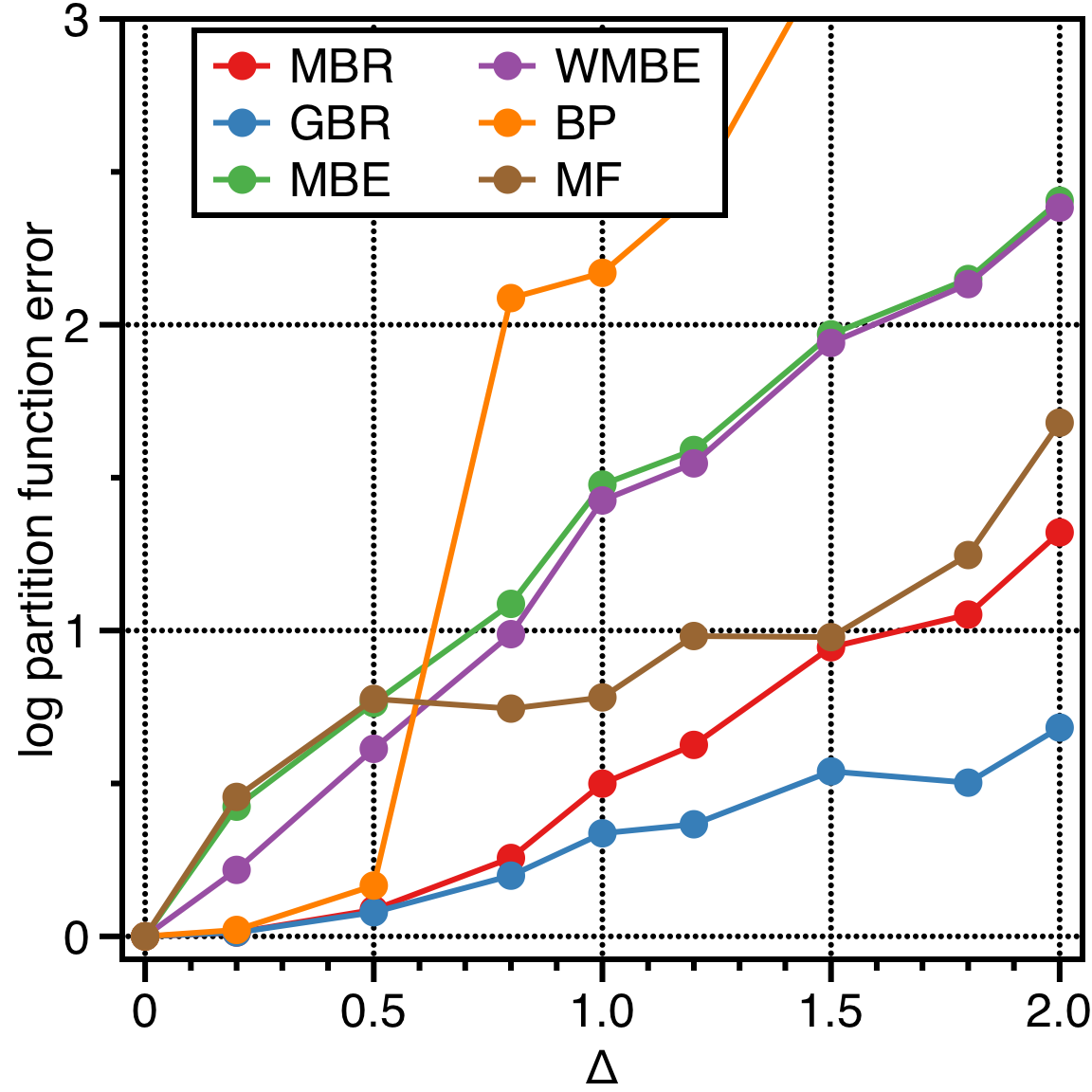}
    \vspace{-0.20in}
    \caption{Complete graph, $|\mathcal{V}|=15$}
    \label{fig:ising_complete_delta}
    \end{subfigure}
    \begin{subfigure}[b]{0.245\linewidth}
    \centering
    \includegraphics[width=0.99\textwidth]{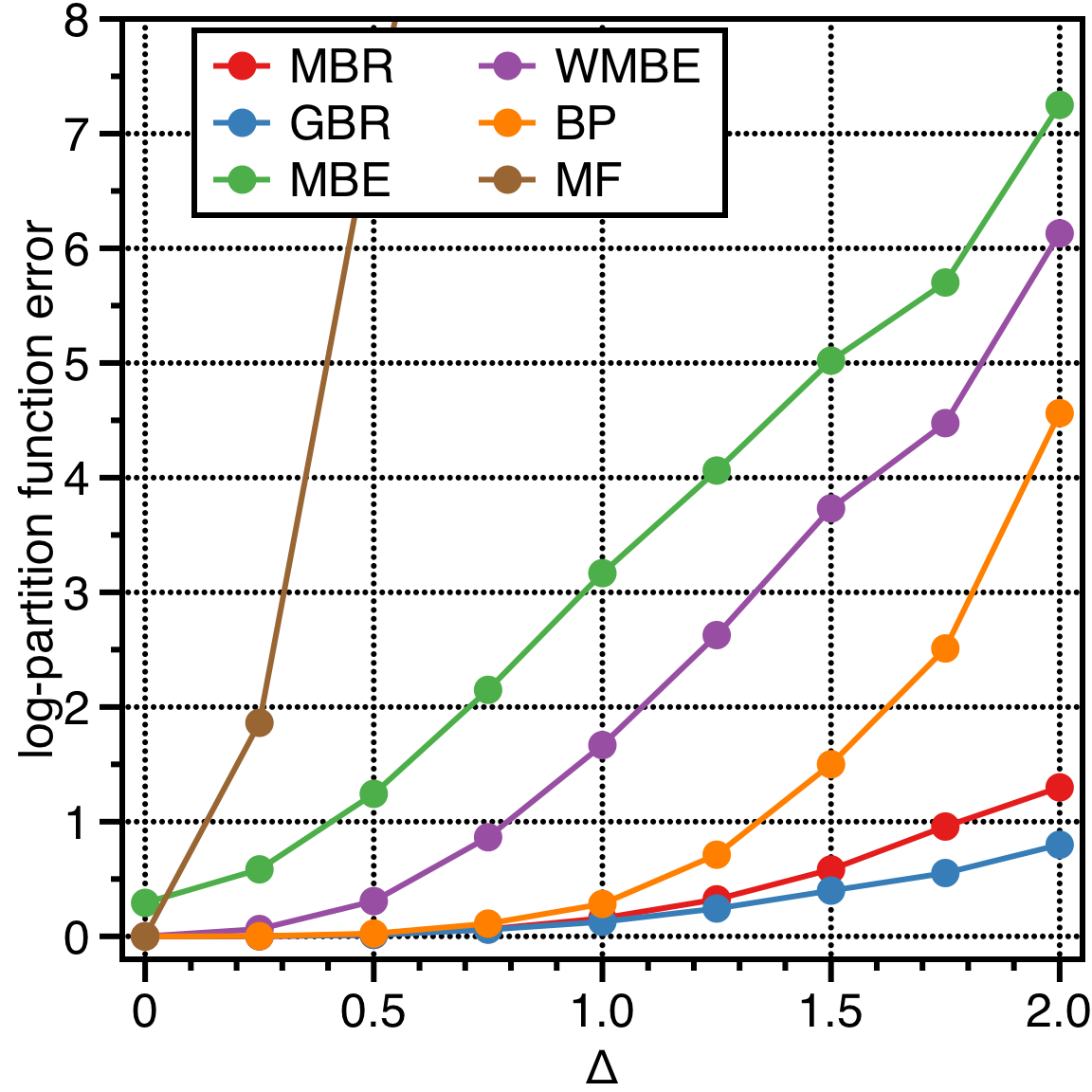}
    \vspace{-0.20in}
    \caption{Grid graph $15 \times 15$}
    \label{fig:ising_grid_delta}
    \end{subfigure}
    \begin{subfigure}[b]{0.245\linewidth}
    \centering
    \includegraphics[width=0.99\textwidth]{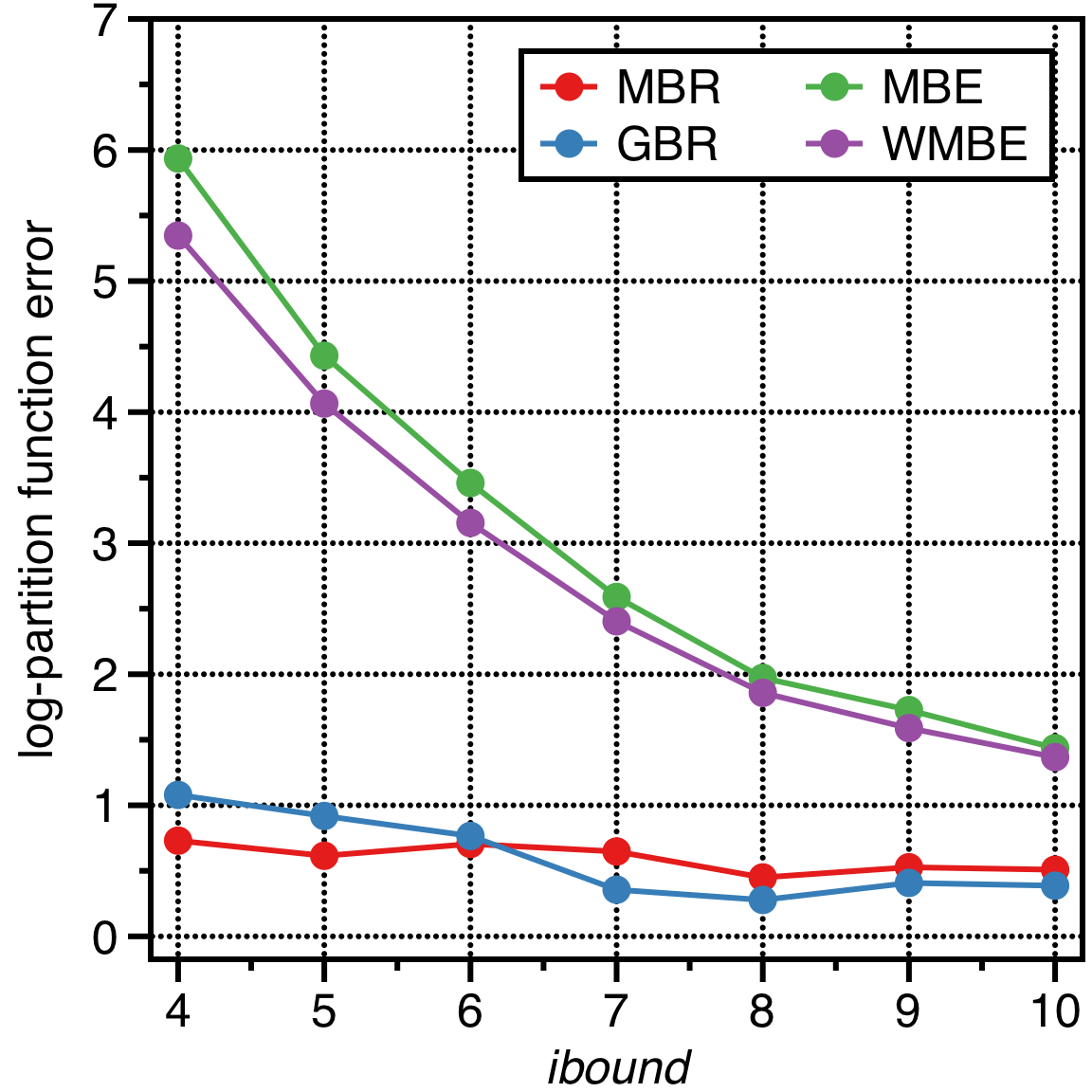}
    \vspace{-0.20in}
    \caption{Complete graph, $|\mathcal{V}|=15$}
    \label{fig:ising_complete_ibound}
    \end{subfigure}
    \begin{subfigure}[b]{0.245\linewidth}
    \centering
    \includegraphics[width=0.99\textwidth]{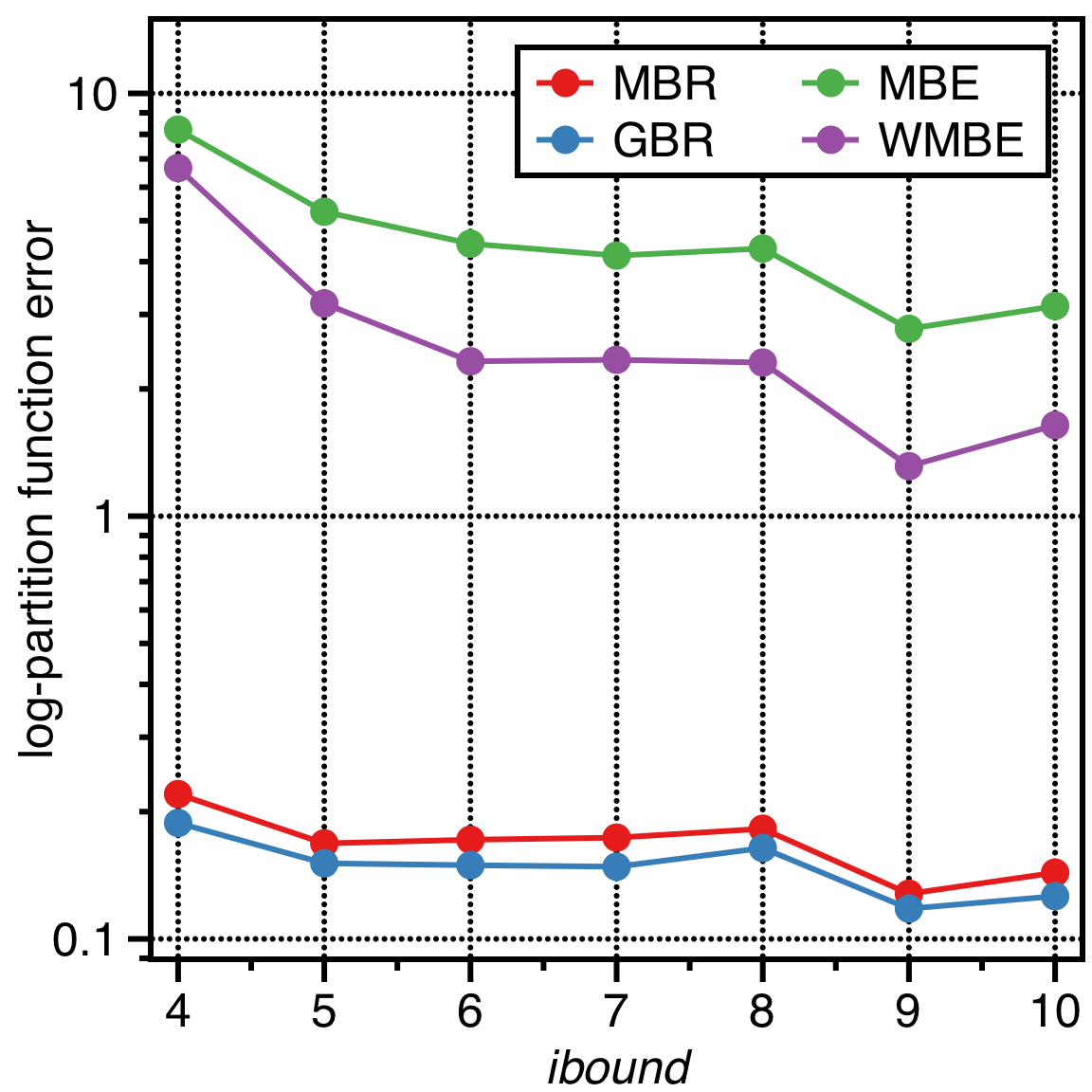}
    \vspace{-0.20in}
    \caption{Grid graph $15 \times 15$}
    \label{fig:ising_grid_ibound}
    \end{subfigure}
    \vspace{0.1in}\\
    \begin{subfigure}[b]{0.249\linewidth}
    \centering
    \includegraphics[width=0.99\textwidth]{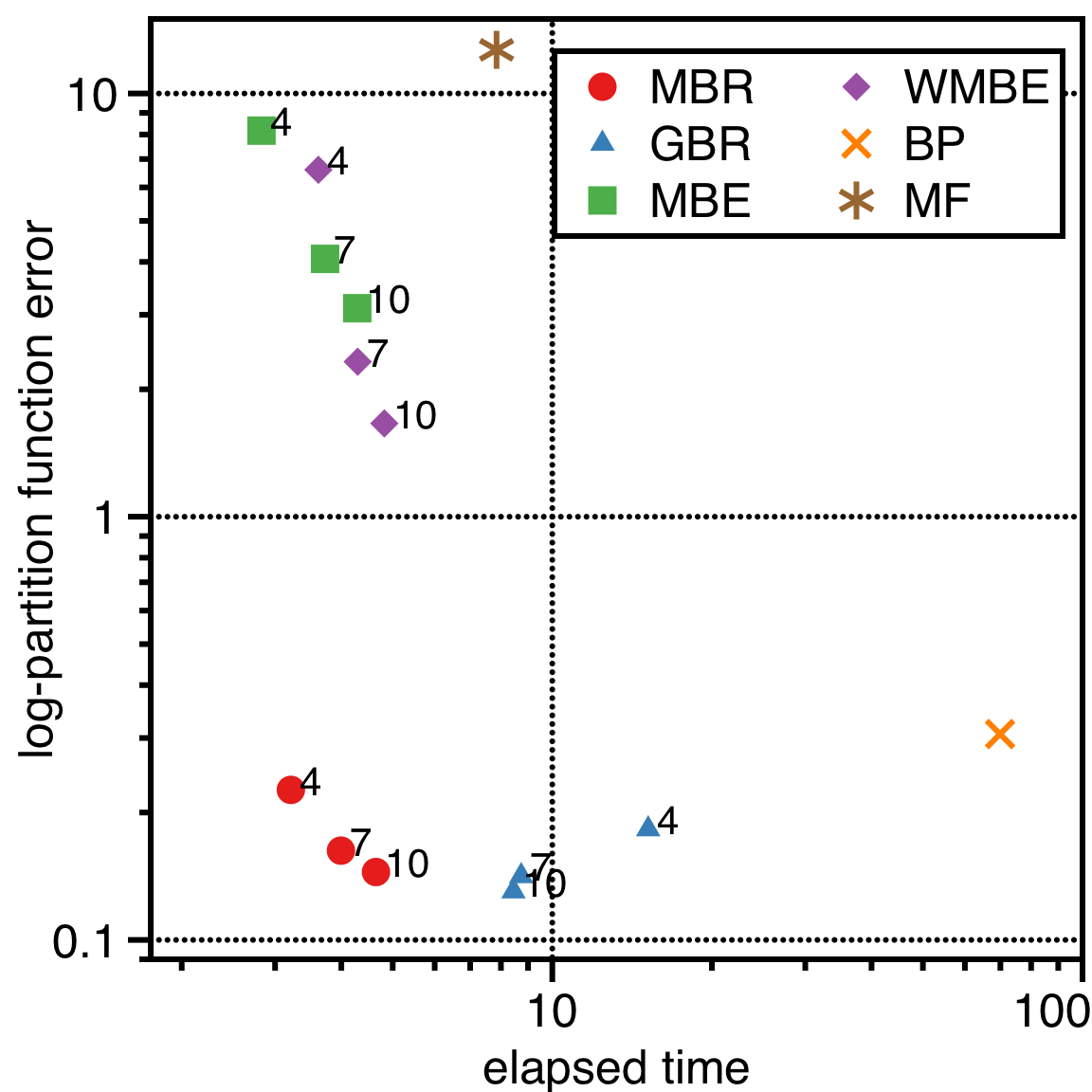}
    \vspace{-0.20in}
    \caption{Grid graph $15 \times 15$}
    \label{fig:ising_grid_time}
    \end{subfigure}
    \begin{subfigure}[b]{0.371\linewidth}
    \centering
    \includegraphics[width=0.99\textwidth]{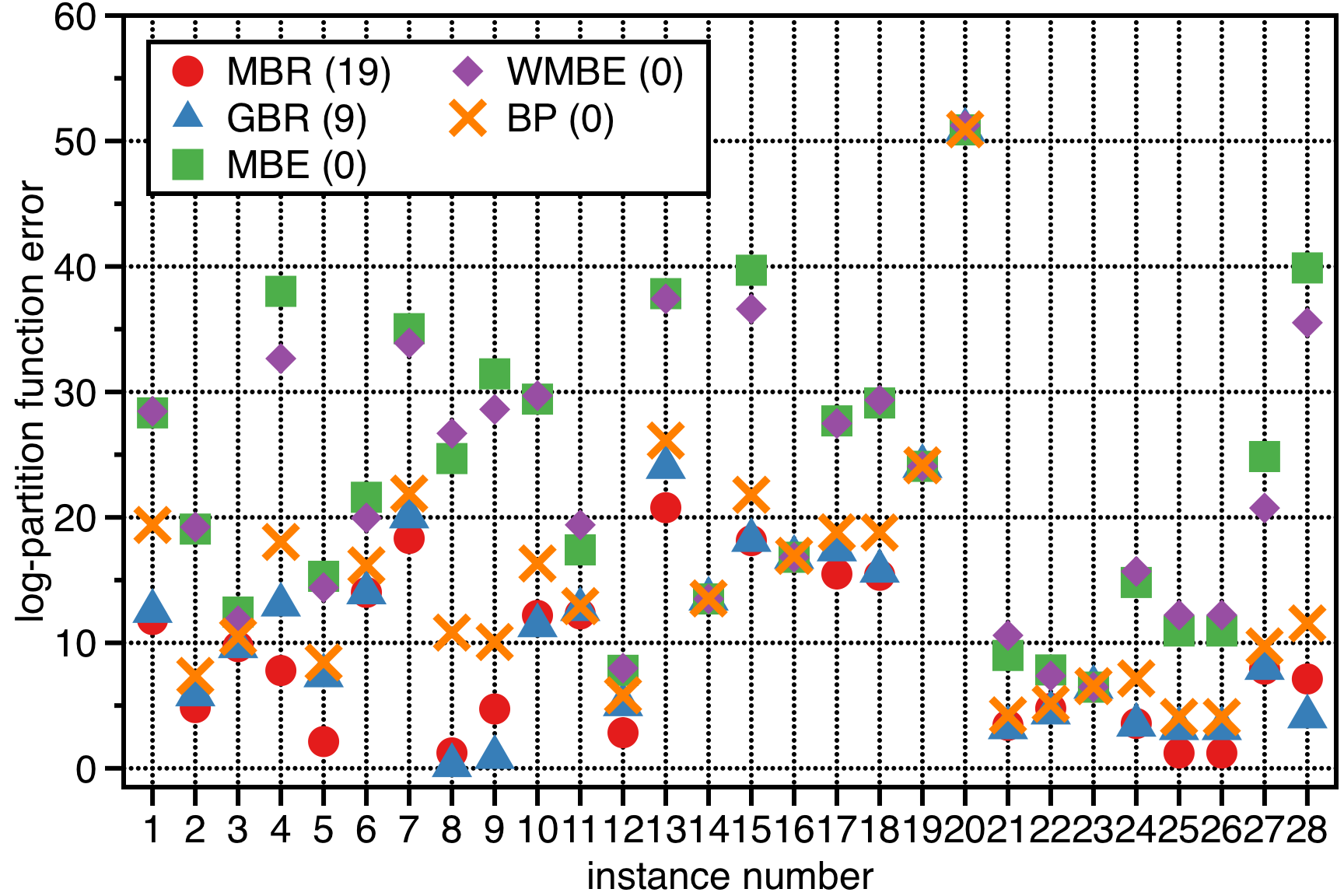}
    \vspace{-0.20in}
    \caption{Promedus dataset}
    \label{fig:promedus}
    \end{subfigure}
    \begin{subfigure}[b]{0.371\linewidth}
    \centering
    \includegraphics[width=0.99\textwidth]{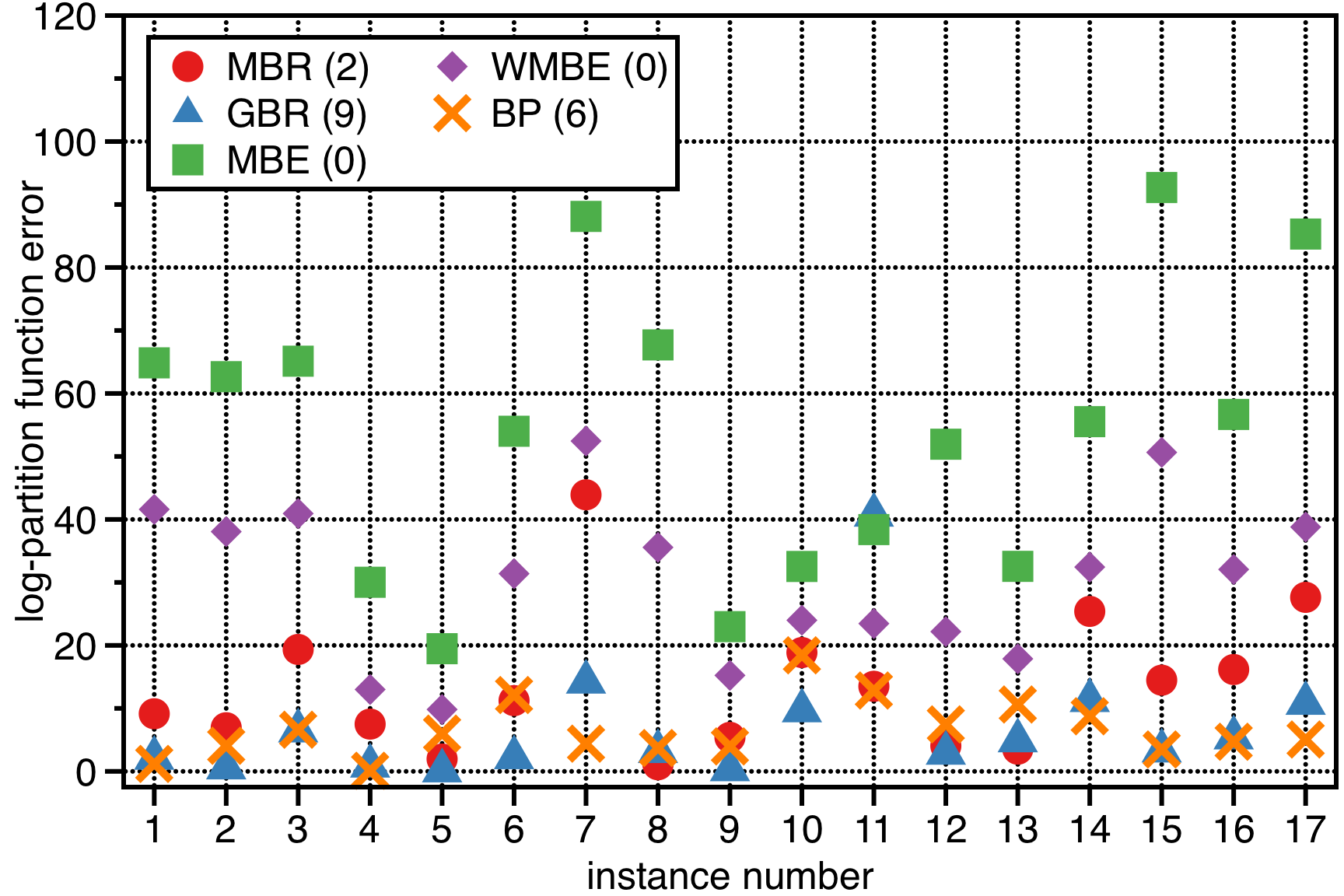}
    \vspace{-0.20in}
    \caption{Linkage dataset}
    \label{fig:linkage}
    \end{subfigure}
    \vspace{-0.3in}
    \caption{
    Performance comparisons under (\subref{fig:ising_complete_delta}, \subref{fig:ising_grid_delta})
    varying interaction strength parameter $\Delta$, and (\subref{fig:ising_complete_ibound}, \subref{fig:ising_grid_ibound}) induced width bound $ibound$. 
    Ising models are either defined on  (\subref{fig:ising_complete_delta}, \subref{fig:ising_complete_ibound})  complete graphs
    with $|\mathcal{V}| = 15$ or (\subref{fig:ising_grid_delta}, 
    \subref{fig:ising_grid_ibound}) grid graphs 
    with $|\mathcal{V}| = 225$. 
    Each plot is averaged over $100$ models.
    (\subref{fig:ising_grid_time}) reports
    the averaged error versus elapsed time while varying $ibound$ in the same setting as (\subref{fig:ising_grid_ibound}),  
    where number labels by points 
    indicate the corresponding $ibound$ used.
    Numbers in brackets, e.g., MBR(19), 
    count the instances of each algorithm to win in 
    terms of approximation accuracy under
    (\subref{fig:promedus}) Promedus  and  (\subref{fig:linkage}) Linkage  datasets. 
    }
    \vspace{-0.1in}
\end{figure*}

In this section, we report experimental results on 
performance of our algorithms for approximating the partition function $Z$.
Experiments were conducted for 
Ising models defined on 
grid-structured and complete graphs
as well as two real-world datasets from the 
UAI 2014 Inference 
Competition \cite{uai}. 
We compare our 
mini-bucket renormalization (MBR) 
and global-bucket renormalization (GBR) scheme 
with 
other mini-bucket algorithms, 
i.e., mini-bucket elimination (MBE) by \citet{dechter2003mini} and 
weighted mini-bucket elimination (WMBE) by \citet{liu2011bounding}.
Here, WMBE used uniform H\"older weights, 
with additional fixed point reparameterization updates 
for improving its approximation, 
as proposed by  
\citealp{liu2011bounding}. 
For all mini-bucket 
algorithms, 
we unified the choice of 
elimination order for 
each instance of GM 
by applying min-fill heuristics \cite{koller2009probabilistic}.
Further, 
we also run the 
popular variational inference algorithms: 
mean-field (MF) and loopy 
belief propagation (BP), 
\footnote{{MF and BP were implemented to 
run for $1000$ iterations at maximum. BP additionally used damping with $0.1$ ratio.}
}
For performance measure, 
we use 
the log-partition function error, 
i.e., 
$|\log_{10}Z - \log_{10}Z_{\text{approx}}|$ 
where $Z_{\text{approx}}$ 
is the approximated 
partition function 
from a respective algorithm.

{\bf Ising models.}
We first consider the most popular binary pairwise GMs, called Ising models \cite{onsager1944crystal}:
    $p(\mathbf{x}) = \frac{1}{Z}\exp\left(\sum_{i\in \mathcal{V}}\phi_{i}x_{i} 
    + \sum_{(i,j)\in \mathcal{E}}\phi_{ij}x_{i}x_{j}\right),$
where $x_{i}\in \{-1, 1\}$. 
In our experiments, we draw 
$\phi_{ij}$ and $\phi_{i}$ 
uniformly from intervals of 
$[-\Delta, \Delta]$ and $[-0.1, 0.1]$ respectively,
where $\Delta$ is 
a parameter controlling the `interaction strength' 
between variables. 
As $\Delta$ grows, 
the inference task 
is typically harder. 
Experiments were 
conducted in two settings: 
complete graphs with $15$ 
variables ($105$ pairwise factors), 
and $15 \times 15$ non-toroidal grid graphs ($225$ variables, $420$ pairwise factors). 
Both 
settings have moderate tree-width, 
enabling exact computation of 
 the partition function 
using BE
with induced widths of 15 and 16, respectively.
We 
vary the interaction 
strength $\Delta$ and the  
induced width bound $ibound$ 
(for mini-bucket algorithms), 
where 
$ibound = 10$ and $\Delta=1.0$ are the default choices.
For each choice of parameters, results are obtained by averaging over 
100 random model instances. 

As shown in Figure \ref{fig:ising_complete_delta}
-\subref{fig:ising_grid_ibound}, 
both MBR and GBR 
perform impressively compared to MF and BP. 
Notably, the relative outperformance 
of our methods compared to the earlier approaches 
increases with 
$\Delta$ (more difficult instances),
and as the bound of induced width gets smaller. 
This suggests that 
our methods 
scale well 
with the size and difficulty of GM. 
In Figure \ref{fig:ising_complete_ibound}, 
where $ibound$ is varied for complete graphs, 
we observe that GBR does not improve over MBR when 
 $ibound$ 
is small, 
but does so after 
$ibound$ grows large. 
This is consistent with our 
expectation that in order for 
GBR to improve over MBR, the 
initial quality of MBR 
should be acceptable.
Our 
experimental setup on Ising grid GMs with 
varying interaction strength is 
identical to that of 
\citet{xue2016variable}, where they 
approximate variable 
elimination in the Fourier domain. 
Comparing results, 
one can observe that our methods
significantly outperform their prior algorithm.

Figure \ref{fig:ising_grid_time} 
reports the trade-off between 
accuracy and elapsed time with 
varying $ibound$. 
Here, we observe that 
MBR is much faster than MBE or WMBE to reach a 
desired accuracy, 
i.e., MBR with $ibound = 4$ 
is faster and more accurate than 
MBE and WMBE 
with $ibound > 4$. 
{
Further, 
we also note that 
increasing $ibound$ for GBR 
does not lead to 
slower running time, 
while the accuracy is improved. 
This is because smaller $ibound$ increases
the numbers of mini-buckets
and corresponding SVD calls. 
}

{\bf UAI datasets.}
We further show results of 
real-world models 
from the UAI 2014 
Inference Competition, 
namely the Promedus (medical diagnosis)  
and Linkage (genetic linkage) 
datasets. 
Specifically, 
there exist $35$ instances of Promedus GMs in 
with the average of $544.75$ binary variables 
and $305.85$ non-singleton 
hyper-edges with 
averaged maximum size
$\max_{\alpha\in\mathcal{E}}|\alpha|=3$. 
In case of Linkage GMs, there exist 
$17$ instances with the average of $949.94$ variables 
with averaged maximum cardinality  $\max_{i\in\mathcal{V}}|\mathcal{X}_{i}| = 4.95$ 
and $727.35$ non-singleton 
hyper-edges with averaged maximum 
size $\max_{\alpha\in\mathcal{E}}|\alpha|=4.47$. 
Again, induced width bounds for mini-bucket algorithms are 
set to $ibound = 10$. 

The experimental results are 
summarized in Figure \ref{fig:promedus} 
and \ref{fig:linkage}. 
Here, results for MF was omitted since 
it is not able to 
run on these instances by its construction. 
First, in Promedus dataset, 
i.e., Figure \ref{fig:promedus}, 
MBR and GBR clearly 
dominates over all other algorithms. 
Even when GBR fails to improve 
MBR, it still outperforms other algorithms. 
Next, in Figure \ref{fig:linkage}, 
one can observe that MBR and GBR 
outperform other algorithms 
in $11/17= 64\%$ of 
the Linkage instances 
and are always as nearly good as the best, i.e., BP, in other instances. 
In particular, our algorithms 
outperform over all 
previous mini-bucket variants 
and expected to outperform BP more significantly
by choosing a larger $ibound$. 

\textbf{Guide for implementation.} 
Based on the experiments, 
we provide 
useful recommendations 
for application of MBR and GBR.
First, we emphasize that 
choosing the right elimination order 
is important for performance. 
Especially, min-fill heuristic 
improved the performance 
of MBR and GBR significantly, 
as with other mini-bucket algorithms. 
Further, whenever 
memory is available, 
running MBR with increased $ibound$ 
typically leads to better 
trade-off between complexity 
and performance 
than running GBR. 
When memory 
is limited, 
GBR is recommended for improving the approximation quality while using the same-order of memory.

\section{Conclusion and Future Work}
We developed a new family of mini-bucket algorithms, MBR and GBR,
inspired by the
tensor network renormalization framework in statistical physics.
The proposed schemes approximate the variable elimination process efficiently by repeating
low-rank projections of mini-buckets.
Extensions to higher-order low-rank projections  \citep{xie2012coarse, evenbly2017algorithms}
might improve performance.
GBR calibrates MBR via minimization of
renormalization error for the partition function explicitly.
A similar optimization was
considered in the so-called second-order renormalization groups \cite{xie2009second, xie2012coarse}.
Hence, there is scope to explore 
potential 
variants of GBR. 
Finally, another direction to generalize MBR and GBR
is to consider larger sizes of buckets to renormalize, e.g., see \cite{evenbly2015tensor, hauru2018renormalization}.
\paragraph{Acknowledgements} AW acknowledges support from the David
MacKay Newton research fellowship at
Darwin College, The Alan Turing Institute
under EPSRC grant EP/N510129/1 \& TU/B/000074,
 and the Leverhulme Trust via the CFI.

\bibliography{references.bib}
\bibliographystyle{icml2018}
\end{document}